\documentclass[12pt]{article}

\usepackage[utf8]{inputenc} 
\usepackage[T1]{fontenc}    
\usepackage{hyperref}       
\usepackage{url}            
\usepackage{booktabs}       
\usepackage{amsfonts}       
\usepackage{nicefrac}       
\usepackage{microtype}      
\usepackage{xcolor}         
\usepackage{enumerate}
\usepackage{natbib}
\usepackage{multirow}
\usepackage{comment}

\usepackage{amsthm}
\usepackage{amsmath}
\usepackage{amssymb}
\usepackage{graphicx}
\usepackage[legalpaper, margin=1in]{geometry}

\newtheorem{definition}{Definition}

\theoremstyle{remark}
\newtheorem{remark}{Remark}

\title{Improving the Validity and Practical Usefulness of AI/ML Evaluations Using an Estimands Framework}

\author{%
  Olivier Binette and Jerome P. Reiter\\
  Duke University
}

\date{\today}

\begin{document}
\maketitle


\begin{abstract}
    Commonly, AI or machine learning (ML) models are evaluated on benchmark datasets. This practice supports innovative methodological research, but benchmark performance can be poorly correlated with performance in real-world applications---a construct validity issue.
    To improve the validity and practical usefulness of evaluations, we propose using an estimands framework adapted from international clinical trials guidelines. This framework provides a systematic structure for inference and reporting in evaluations, emphasizing the importance of a well-defined estimation target. We illustrate our proposal on examples of commonly used evaluation methodologies---involving cross-validation, clustering evaluation, and LLM benchmarking---that can lead to incorrect rankings of competing models (rank reversals) with high probability, even when performance differences are large.  We demonstrate how the estimands framework can help uncover underlying issues, their causes, and potential solutions. Ultimately, we believe this framework can improve the validity of evaluations through better-aligned inference, and help decision-makers and model users interpret reported results more effectively.
\end{abstract}


\section{Introduction}

Evaluating AI or machine learning (ML) models is critical at all stages of ML projects, influencing both development and deployment phases \citep{cohen1989toward, reich1999evaluating, Schelter2015}. It facilitates comparisons among algorithms, guides feature selection and training, and allows for iterative refinements while ensuring robust performance in production settings.

Commonly, models are evaluated by measuring performance on benchmark or test datasets \citep{liao2021we}. The practice has many limitations despite being a key contributor to methodological progress in the field \citep{dehghani2021benchmark}. In many disciplines, benchmark performance metrics often do not generalize well to real-world capability \citep{liao2021we, Wang2022}. \cite{ferrari2019we} and \cite{hutson2020core} documented ``phantom progress,'' where inappropriate use of benchmark datasets and baseline methods leads to misleading performance estimates and an illusion of progress. \cite{oakden2020hidden} showed how ``hidden stratification,'' where meaningful subgroups are not identified in benchmark datasets, can lead to hidden failure modes that performance metrics fail to represent. More broadly, \cite{hutchinson2022evaluation} observed that the ``idealized breadth of evaluation concerns'' is not reflected in common benchmark-based evaluation practices. These types of issues are sometimes referred to as construct validity issues, i.e., a misalignment between theoretical goals and practical measurement or inferential methods \citep{sjoberg2022construct, biderman2024lessons}.

To help address these issues, we propose adapting the \textit{estimands framework} from international clinical trials guidelines \citep{ICH2019, phillips2021estimands} to ML evaluation. The goal of the framework is to better align evaluation objectives with the design of evaluations (e.g., how to acquire data and what measurements to make) and the data analysis (e.g., how to summarize results and how to make inferences). It achieves this by emphasizing the importance of having well-defined targets of estimation, the estimands, to enable aligned and efficient evaluations. Without well-defined estimands, evaluation stops at taking measurements and cannot make meaningful generalizations or inferences, or cannot clearly report results that a broad community of users.


\begin{quote}
    \textit{``Incorrect choice of estimand and unclear definitions for estimands lead to problems in relation to trial design, conduct and analysis and introduce potential for inconsistencies in inference and decision making.'' \citep{ich2014final}}
\end{quote}

\begin{figure}[!ht]
    \centering
    \includegraphics{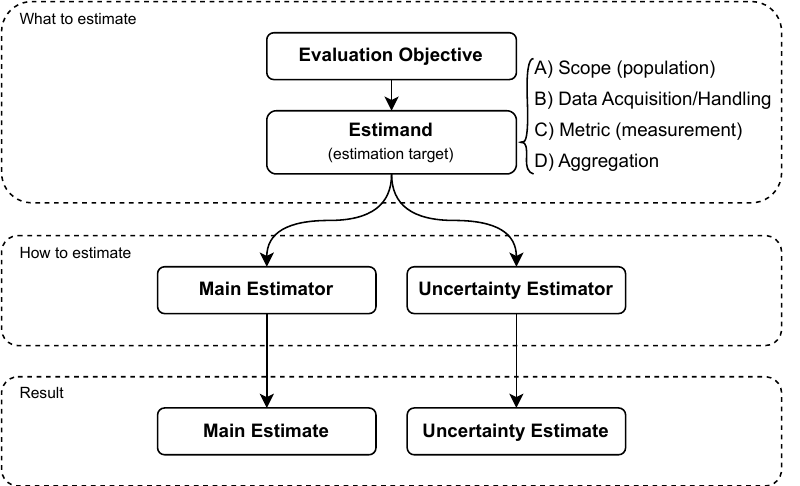}
    \caption{Estimands framework adapted from \cite{ICH2019} for ML model evaluation, as described in Section \ref{sec:estimand-framework}. An evaluation objective is translated to an estimand. An estimand is characterized by (A) a metric or choice of measurement, (B) a specific scope (a population) to contextualize the metric, (C) a data acquisition strategy (including how missing data, data annotation inconsistencies, and other data issues are handled), and (D) an aggregation/summarization of the metric values over the given scope/population. Next, a main estimator is chosen to provide a sufficiently accurate estimate at minimal cost. Uncertainty regarding the estimation procedure can be separately or jointly estimated, accounting for sensitivity to the choice of the main estimator and its underlying assumptions.}
    \label{fig:estimand-framework}
\end{figure}

The estimands framework formalizes essential best practices for experimental design, providing key steps to accurately describe the estimation target (the estimand) and emphasizing the subtler considerations that contribute to a meaningful definition. It is quite simple and straightforward, but nonetheless an important reminder and standardized structure for key components that must be considered in applications.
%
Figure \ref{fig:estimand-framework} provides an overview of the framework adapted to ML evaluation, illustrating the components of an estimand and its relationship with an evaluation objective and data analysis. More details are given in Section \ref{sec:estimand-framework}.

To support our proposal, we consider three examples that demonstrate failures of commonly used evaluation methodologies, and how the estimands framework reveals causes and solutions. The examples are related to a fundamental evaluation problem: the accurate ranking of ML models according to a chosen dimension of performance. We define a performance rank reversal as occurring when a model is wrongly deemed superior to another, despite the opposite being true (see Section \ref{sec:defining_rank_reversals}). 
%

Our examples show that rank reversals can occur using common evaluation methodologies in simple applications, despite substantial performance differences between models. They are based on practices commonly found in the literature, but are simplified to demonstrate that problems can arise even in the straightforward scenarios. The three examples are:
\begin{description}
    \item[Cross-Validation Example (Section \ref{sec:cross-validation-rank-reversal}):] Unbiased cross-validation estimators \citep{stone1974cross, bates2023cross} are widely used for model selection. We show in a simple regression example how cross-validation can lead to the selection of the worse model with high probability.
    \item[Clustering Evaluation Example (Section \ref{sec:clustering-example}):] Evaluating clustering models for entity resolution applications \citep{Christophides2019}, such as identity clustering based on face images \citep{Shi2018face}, often relies on computing an F-score on a small benchmark dataset \citep{Shi2018face, Yin2020}. We show how the resulting F-score can be biased and unreliable for ranking models.
    \item[LLM Benchmarking Example (Section \ref{sec:LLM-rank-reversal}):] We show how the composition of LLM benchmark datasets \citep{srivastava2022beyond} along unmeasured dimensions can affect the relative performance of LLMs, making it difficult for rankings to generalize.
\end{description}

We apply the framework to each example, in order to show its practical use in reviewing and developing evaluations. We also discuss some of the subtler issues involved in the definition of an estimand. Specifically, the application to the cross-validation example shows the importance of considering context and population for valid inferences. We use the clustering evaluation example to emphasize the impact of data acquisition issues on the definition of an estimand, and we use the LLM benchmarking example to discuss the potential of multi-criteria decision-making methods. 


In summary, by using the estimands framework as scaffolding, we can ensure that ML evaluations are well aligned with key goals, that they produce valid inferences, and that their results are meaningful for applications and model users. 

The rest of the paper is organized as follows. In Section \ref{sec:background_evaluation}, we provide background on ML evaluation and the approach of our paper by describing the importance of measurement, inference, and reporting in evaluations. Section \ref{sec:three-examples} describes our three examples. We introduce our examples before the estimands framework to show how certain evaluation problems can be unexpected or surprising when they are not properly contextualized through the framework. Section \ref{sec:estimand-framework} introduces the estimands framework and applies it to each example. Section \ref{sec:discussion_estimands_framework} summarizes our findings and discusses broader potential for the estimands framework to improve ML evaluations.


%

\section{Background} \label{sec:background_evaluation}

In Section \ref{sec:definitions-related-work}, we define ``models'' and ``evaluation,'' and discuss common ML evaluation practices and their goals. In Section \ref{sec:dedicated-discipline}, we discuss our approach to ML evaluation.

\subsection{Definitions and Related Work}\label{sec:definitions-related-work}

We use \textbf{model} as an umbrella term for trained and untrained models, ML algorithms, and ML/AI systems. The scope is broad since we focus on statistical evaluation rather than any particular ML subfield. The statistical evaluation principles we discuss are widely used in applied statistics and other domains, such as clinical trials biostatistics. Therefore, we believe they are also useful for a wide range of ML applications.

We define \textbf{evaluation} as a study with the goal of making \textit{value judgments} to guide \textit{action, decision, or change}. See \cite{wanzer2021evaluation} for relevant discussion. We emphasize the scientific components of evaluation, and its goal of providing judgments that have practical consequences. The importance of judgemental evaluation is emphasized in \cite{mathison2005evaluation}, and the importance of action-oriented outcomes is emphasized in \cite{tong1987performance}. For example, clinical trials aim to determine the efficacy, safety, and other characteristics of medical treatments, with a direct impact on clinical practice. Evaluation may be focused on developing cost-effective methods. For instance, adaptive designs are developed to reduce costs, improve accuracy, and improve patient outcomes in clinical trials.

There are three core components to evaluation: \textbf{measurement}, \textbf{testing/inference}, and \textbf{reporting}. Measurement captures a given characteristic of an object or state of the world. For example, in LLM evaluation, measurements are the scoring of responses on evaluation items. A large literature investigates techniques to efficiently and reliably score responses using human judges or automated methods \citep{liu2023gpteval, zhang2023wider, zheng2024judging}. Testing/inference involves checking assumptions and expectations, often probabilistically, to determine if we are likely right or wrong. In ML evaluation, inference might translate measurements from a training dataset to an expected generalization error. Statistical testing can assess the significance of performance differences between models \citep{dehghani2021benchmark}. Reporting involves summarizing and communicating evaluation results, addressing the needs of its consumers. It bridges scientific insights and real-world change, requiring sufficient effort in summarization, communication, and analysis to support actions or change.

Too often, ML evaluations stop at measurement, only computing scores on a benchmark dataset \citep{post2018call, dehghani2021benchmark, colombo2022best, srivastava2022beyond}. In these cases, there is often no uncertainty regarding the target of estimation and no direct consideration of how performance might generalize beyond the benchmark.

Even when inferences are made, say by estimating generalization performance through cross-validation, the process may not align with evaluation objectives. For instance, it is known that cross-validation estimators are often only weakly correlated, or even negatively correlated, with the generalization performance of a given model \citep{hastie2009elements, bates2023cross}. Whether or not a cross-validation estimate is representative of the generalization performance of a given trained model must be checked. This example is discussed in more detail in Section \ref{sec:cross-validation-answer}.

On the reporting front, standardized approaches like data cards and model cards are widely used \citep{mitchell2019model, pushkarna2022data}. But interpreting benchmarking results and performance evaluations can be difficult. Complex or unmeasured characteristics of a benchmark dataset impede understanding of what it represents and how results translate into practice \citep{dehghani2021benchmark}. Another problem is the lack of reporting of disaggregated metrics or item-level performance \citep{Burnell2023rethink}. Furthermore, relatively little attention has been given to aggregating or reporting scores from multiple tasks \citep{colombo2022best}. Even with a clear evaluation objective, determining the adequacy of the evaluation methodology can be challenging. As shown in our examples (Section \ref{sec:three-examples}), many evaluation methodologies lead to rank reversals with high probability.



\subsection{ML Evaluation as a Discipline}\label{sec:dedicated-discipline}

We treat ML evaluation as a dedicated discipline, focusing on applied performance estimation rather than model development. While evaluation is often tightly coupled with model development, it should also be considered as a separate topic to (1) ensure the robustness and efficiency of evaluations, (2) avoid conflicts of interest or misaligned incentives in high-risk applications, as emphasized in \cite{SR1107}, and (3) ensure that evaluation results are useful to decision-makers or model users. This separation is particularly important given the increased usage of general-purpose and pre-trained ML models across a wide range of software systems. In this context, evaluation is a separate activity closely tied to an application area rather than the development or refinement of a model. The large number of ML models used in many organizations motivates the development of efficient methodologies and systematic evaluation processes applicable across multiple domains.

\section{Three Examples Leading to Rank Reversals}\label{sec:three-examples}

In this section, we define rank reversals and describe three examples where commonly used evaluation methods have high probability of rank reversal. 

The examples provide motivation for the estimands framework, which is introduced next in Section \ref{sec:estimand-framework}. We delay explaining the cause for evaluation problems until Section \ref{sec:estimand-framework}, where we revisit each example to show how the framework provides the necessary scaffolding to follow best practices and resolve issues. Our examples are intentionally simplified to highlight common evaluation problems, not best practices. 

\subsection{Defining Performance Rank Reversals}\label{sec:defining_rank_reversals}

Consider two models, $A$ and $B$, that we want to rank by a performance metric $\varphi$ such as generalization accuracy. These models may be pre-trained and fixed, or viewed as random to account for variability in the training process or training data. We denote by $\varphi(A)$ and $\varphi(B)$ the scores of the two models under $\varphi$. A statistical evaluation methodology provides two corresponding estimators $\hat \varphi(A)$ and $\hat \varphi(B)$. Although estimating $\varphi$ directly can be challenging in some situations, the ranking of alternatives generated by $\hat \varphi$ should reflect the ranking produced by $\varphi$, with sufficient probability. If not, the estimators are said to cause a rank reversal. 

We formalize the general notion of rank reversals in Definition \ref{def:rank_reversal} below.

\begin{definition}[Rank Reversals]\label{def:rank_reversal}
    Let $\varphi$ be a performance metric of interest, and let $A$, $B$ be two given models. The probability of rank reversal for the estimators $\hat \varphi(A)$ and $\hat \varphi(B)$ is defined as
    \begin{equation}\label{eq:probability_correct}
        \mathbb{P}\Big( \text{rank}\left\{\hat \varphi(A), \hat \varphi(B)\right\} \neq \text{rank}\left\{\varphi(A), \varphi(B)\right \} \Big),
    \end{equation}
    where the probability taken with respect to the joint distribution of the estimators and the models $A$ and $B$.
\end{definition}

Note that the underlying probability space varies by context. In our cross-validation example of Section \ref{sec:cross-validation-rank-reversal}, we analyze properties of estimators from the perspective of hypothetical resampling of a training dataset and the following model training. In our clustering evaluation example of Section \ref{sec:clustering-example}, we consider two fixed, pre-trained models, and randomness is introduced by labeling a sample of the data. In our LLM example of Section \ref{sec:LLM-rank-reversal}, everything is fixed and the probability of a rank reversal is either $0$ or $1$.

Rank reversals have significant consequences and costs in applications, beyond the direct impact of performance loss. For example, in the early stages of model development, rank reversals can lead to discarding useful features or modalities, thereby hindering future performance \citep{Wang2022}. These effects can compound throughout model development, leading to high costs.

\subsection{Rank Reversals With Cross-Validation}\label{sec:cross-validation-rank-reversal}

\subsubsection{Background on Cross-Validation}

Consider a supervised learning problem with independent and identically distributed data points $(X_i, Y_i) \in \mathcal{X} \times \mathcal{Y}$ for $i \in \mathbb{N}$. A training algorithm $\mathcal{A}$ maps a set of examples $\mathcal{D}$ to a model represented by a function $\hat f_{\mathcal{D}} : \mathcal{X} \rightarrow \mathcal{Y}$. Given a loss function $\ell(\hat y, y) \geq 0$ and $n$ examples $\mathcal{D}_{n} = \{(X_1, Y_i)\}_{i=1}^n$, the training algorithm aims to learn a function $\hat f_{\mathcal{D}_n}$ that minimizes the expected generalization error
\begin{equation}\label{eq:gen_error}
    \text{Err}_{\text{gen}}(\hat f_{\mathcal{D}_n}) = \mathbb{E}\left[ \ell\left(\hat f_{\mathcal{D}_n}(X_{n+1}), Y_{n+1} \right) \mid \mathcal{D}_{n}\right].
\end{equation}
This generalization error cannot be computed exactly without knowledge of the distribution of $(Y_i, X_i)$. However, it is commonly estimated using cross-validation. For example, with leave-one-out cross-validation, we consider a point $(X, Y) \in \mathcal{D}_n$ sampled at random, a function $\hat f_{\mathcal{D}^*}$ learned from $\mathcal{D}^{*} = \mathcal{D}_n\backslash \{(X, Y)\}$, and we compute
\begin{equation}
    \text{CV}_{\text{loo}}(\hat f_{\mathcal{D}_n}) = \mathbb{E}\left[\ell\left(\hat f_{\mathcal{D}^*}(X), Y\right)\mid \mathcal{D}_{n}\right] = \frac{1}{n}\sum_{i=1}^{n} \ell\left(\hat f_{\mathcal{D}_n \backslash \{(X_i, Y_i)\}}(X_i), Y_i)\right).
\end{equation}
The cross-validation estimator is considered ``unbiased'' in the sense that
\begin{equation}
    \mathbb{E}\left [\text{Err}_{\text{gen}}(\hat f_{\mathcal{D}_n}) - \text{CV}_{\text{loo}}(\hat f_{\mathcal{D}_n}) \right ] = 0.
\end{equation}

\begin{remark}
It is well known that the cross-validation estimator $\text{CV}_{\text{loo}}(\hat f_{\mathcal{D}_n})$ is better at estimating the unconditional generalization error $ \mathbb{E}\left [\text{Err}_{\text{gen}}(\hat f_{\mathcal{D}_n})\right]$ than the conditional generalization error $\text{Err}_{\text{gen}}(\hat f_{\mathcal{D}_n})$ \citep{hastie2009elements, bates2023cross}. However, in practice, the conditional generalization error is the target metric of interest since we care about the performance of a realized model. The key observations of Section \ref{sec:california_example} below do not change based on the choice of conditional or unconditional generalization error as a target metric. Rank reversals occur with high probability for both choices of target metrics.
\end{remark}

\subsubsection{Example: The California Housing Dataset}\label{sec:california_example}

Unfortunately, cross-validation estimators can lead to a high probability of rank reversal, even in simple examples with large performance differences between two models.

We illustrate this problem in a simple, standard example. Consider the California Housing Dataset, which contains information on house attributes in $N = 20,640$ California \textit{block groups} from the 1990 U.S. Census. For a given Census block group $i$, the response variable $Y_i$ is the Census block group's median house value. The feature vector $X_i$ has $8$ numerical dimensions. The goal is to learn a predictive function for the median house value that minimizes the generalization performance, using a mean squared error loss function. We consider a training dataset $\mathcal{D}_n$ of $n=2,000$ examples, randomly selected with replacement from the set of California Census block groups, and two training algorithms: (A) a simple linear regression, and (B) a decision tree model with a maximum depth of $5$, using default scikit-learn parameters for their implementation. These training algorithms were chosen for simplicity, rather than for accuracy or effectiveness. The true difference in performance between the two models is rather substantial. The root mean squared error of the linear model is around $\$200,000$, versus $\$74,000$ for the decision tree (the median house price ranges from $\$30,000$ to $\$500,000$).

\begin{table}[htb!]
\centering
\begin{tabular}{rcc}
\toprule
    & \textbf{Linear Model} & \textbf{Decision Tree} \\ 
\midrule
    \textbf{Avg. of generalization MSE $(\varphi)$} &  4.02 & 0.553 \\
    \textbf{Avg. of CV estimates} & 4.00 & 0.553 \\
    \cmidrule(lr){1-3}
    \textbf{Probability of rank reversal} & \multicolumn{2}{c}{75.3\%}\\ 
\bottomrule
\end{tabular}
\caption{\label{tab:cv-results} Summary of our cross-validation experiment. Decision trees are generally better at predicting median house price than linear models as seen by their lower average mean squared error (MSE). Furthermore, the average of the cross-validation (CV) estimates are close to the average MSE, showing that that CV estimates are nearly unbiased. However, in the majority of the replications (around $75\%$ of them), the worst-performing model is selected.}
\end{table}

Next, we compute the rank reversal probability for the leave-one-out cross-validation estimator $\text{CV}_{\text{loo}}$ with respect to models resulting from sampling a training dataset $\mathcal{D}_n$ and the following target performance metric:
\begin{itemize}
    \item[$\varphi$:] the expected generalization error $\text{Err}_{\text{gen}}$ defined in \eqref{eq:gen_error}, computed over all $N=20,640$ California Census block groups.
\end{itemize}
We estimate the probability of rank reversal using $50,000$ replications of sampling a training dataset of size $n=2,000$, fitting each model, and comparing their leave-one-out cross-validation performance estimate to their true generalization performance. 

The results of the experiment are summarized in Table \ref{tab:cv-results}, showing that the cross-validation estimator leads to a rank reversal with a probability of $75.3\%$.\footnote{When choosing the unconditional generalization error $\mathbb{E}\left [\text{Err}_{\text{gen}}(\hat f_{\mathcal{D}_n})\right]$ as the target metric rather than the conditional generalization error, the probability of rank reversal is slightly lower at $66\%$.}
Given this high probability of rank reversal, the cross-validation is not appropriate for choosing between linear and decision tree models in this application.


\subsection{Rank Reversals in Clustering Evaluation}\label{sec:clustering-example}

\subsubsection{Background on Clustering Evaluation}

Consider a clustering problem, where each element $r$ of a set $\mathcal{R}$ belongs to a cluster $c(r) \subset \mathcal{R}$, and where the resulting set of clusters $\mathcal{C}$ partitions $\mathcal{R}$ (each element $r$ belongs to a single cluster $c(r) \in \mathcal{C}$). The goal is to predict $\hat c(r) \subset \mathcal{R}$ for the cluster to which each element $r$ belongs, under the same constraint that the resulting set of predicted clusters $\hat{\mathcal{C}}$ partitions $\mathcal{R}$.

This can be equivalently formulated as a classification problem, where the goal is to predict whether two elements $r, r' \in \mathcal{R}$ belong to the same cluster under a transitivity constraint. Thus, classification evaluation metrics such as precision $P$, recall $R$, and the F-score $F$ are commonly used to evaluate clustering models. Denoting by $\mathcal{P}$ the set of pairs of elements predicted to belong to the same cluster and by $\mathcal{T}$ the set of pairs that truly belong to the same cluster, these metrics are defined as
\begin{equation}
    P = \frac{\lvert \mathcal{T} \cap \mathcal{P} \rvert}{\lvert\mathcal{P} \rvert}, \quad R = \frac{\lvert \mathcal{T} \cap \mathcal{P} \rvert}{\lvert\mathcal{T} \rvert}, \quad F = \left( \frac{P^{-1} + R^{-1}}{2} \right)^{-1}.
\end{equation}

Since the true clustering is unknown, these metrics can only be computed for labeled benchmark datasets or labeled subsets of the data. A common practice in the literature is to compare clustering models based on the F-score computed on a benchmark dataset \citep{xie2013overlapping, Yin2020, Wang2022}. In the context of our framework defined in Section \ref{sec:defining_rank_reversals}, our target metric is $\varphi = F$, with the estimator $\hat \varphi$ being F-score computed on a benchmark dataset or a labeled data subset, and the clustering models considered to be pre-trained.

\subsubsection{Example: Identity Clustering Based on Face Images}

Unfortunately, the F-score tends to degrade with dataset size in a nonlinear manner, depending on specific characteristics of the clustering models. Thus, a clustering model $A$ may outperform model $B$ on small subsets of the data, but model $B$ may outperform model $A$ on the full dataset because it is less affected by performance degradation as dataset size increases \citep{binette2023estimating}. If performance on a labeled data subset is used to select a model for clustering the full data, this causes a rank reversal: model $A$ is chosen, even though model $B$ performs better on the full dataset.

To illustrate this problem, we consider an identity clustering task related to facial recognition. Specifically, we use the Olivetti Faces dataset\footnote{The Olivetti Faces dataset was created at AT\&T Lab Cambridge and obtained online from scikit-learn 1.3.} which contains 400 face pictures of 40 individuals, with 10 images per person. The goal is to cluster the images by individual identity. To solve this task, we use k-means clustering on pre-trained FaceNet embeddings \citep{schroff2015facenet, Timesler}. Model $A$ uses $k=30$ clusters and model $B$ uses $k=60$ clusters. The true F-score, which would be unknown in practice, can only be computed using the cluster membership of all 400 face pictures. It is estimated by computing the $F$-score on a benchmark dataset of $10$ randomly selected individuals for whom the true clusters have been resolved. Table \ref{tab:clustering_results} shows the $F$-scores of the two models on the full dataset, the average $F$-score estimates, and the probability of rank reversal in $20,000$ simulations of sampling 10 individuals to estimate the F-score. There is a $66\%$ probability of rank reversal: the F-score estimator selects model $B$ $66\%$ of the time, even though model $A$ performs better on the full dataset. 

\begin{table}[htb!]
\centering
\begin{tabular}{rcc}
\toprule
    & \textbf{Model $A$} & \textbf{Model $B$} \\ 
\midrule
    \textbf{True F-score} &  \textbf{0.87} & 0.73 \\
    \textbf{Avg. of F-score estimates} & 0.87 & \textbf{0.90} \\
    \cmidrule(lr){1-3}
    \textbf{Probability of rank reversal} & \multicolumn{2}{c}{66\%} \\ 
\bottomrule
\end{tabular}
\caption{\label{tab:clustering_results} Summary of the clustering evaluation experiment. Model $A$ is better than model $B$ in terms of F-score computed on the entire dataset (0.87 versus 0.73).
However, the F-score estimator (computing the F-score on a random subset of 10 labeled clusters) is highly biased, leading to the worst model being selected around $66\%$ of the time.}
\end{table}

\subsection{Rank Reversals in LLM Benchmarking}\label{sec:LLM-rank-reversal}

\subsubsection{Background on Benchmarking Large Language Models}

Methodological research on large language models (LLMs), like in other ML fields, relies on common benchmark datasets to evaluate models and track progress \citep{liao2021we, dehghani2021benchmark, lin2021truthfulqa, srivastava2022beyond, colombo2022best, zhou2023don, chang2023survey, guo2023evaluating}. Benchmark datasets are organized by task, topic, and other characteristics, with corresponding evaluation items. For example, the Format-Following benchmark dataset \citep{xia2024fofo} tasks an LLM to follow formatting guidelines specified in a prompt. This benchmark dataset contains $494$ evaluation items across $10$ application domains, $50$ subdomains, and $248$ format types. Each evaluation item instructs the LLM to format data in a specified way. Success in format-following can be assessed by human annotators and/or a judging model.

The objective is to determine how well a given LLM should perform on tasks similar to the ones in the benchmark dataset. That is, the target metric $\varphi$ is the generalization performance for a given task description, and the estimator $\hat \varphi$ is observed performance on the task's benchmark.

There are two main challenges in evaluating LLMs. First, the open-endedness of expected answers. There is not always a single correct answer to a given evaluation item. In such cases, correctness is determined by a separate judge. Second, LLMs are evaluated for broad humanlike capabilities rather than for precisely defined quantitative objective.

In practice, these challenges are addressed by comparing LLMs to humans, effectively anthropomorphizing LLM capabilities \citep{chollet2019measure, chang2023survey}. LLMs are evaluated similarly to humans and are compared to humans to contextualize their performance \citep{srivastava2022beyond, chang2023survey}. Consequently, many implicit assumptions related to educational and psychological measurement are applied to LLMs to facilitate the interpretation of results. For example, statistical models used to quantify human performance typically assume a latent trait, such as 'mathematical reasoning ability,' estimated based on item responses. The scale of that latent variable may be ignored, as long as a well-defined population of reference can be used for z-scores. Through these latent traits, we can generalize to expected performance on similar tasks. For example, a human or LLM that performs better than others on a mathematical exam might be expected to perform better than others on similar mathematical tasks.

In this line of thinking, \cite{hernandez2017evaluation, chollet2019measure, martinez2019item, wang2023evaluating} proposed the use of psychometric methods for evaluating ML systems. For instance, item response theory \citep{cai2016item, lalor-etal-2024-item} is one way to formalize the estimation of latent traits from evaluation items. Note that there are limitations to traditional psychometric-based approaches, such as the reliance on a reference population, for which promising alternatives have been proposed \citep{hernandez2022training, burnell2022not, burden2023inferring, mlayoutstutorial2024}.
In common practice, the use of LLM benchmarks is typically quite simple, only measuring the average performance on evaluation items. These performance averages are sometimes disaggregated by topic or by other characteristics.

\subsubsection{Example: The Format-Following Benchmark Dataset}

Unfortunately, current LLMs do not always behave like humans (see e.g., \cite{efrat2211lmentry, wang2023evaluating}), leading to unexpected performance rank reversals in some applications. LLM rankings for a given task's benchmark dataset may not be maintained when considering other similar tasks, due to unmeasured confounders that would not be expected to impact human rankings to the same extent. 

For instance, LLM rankings on the Format-Following benchmark are not always stable across different difficulty levels (see table \ref{tab:llm-rankings}). An LLM may outperform another on difficult questions but perform worse on easy questions, or vice versa. Similar rank reversals can be observed in \cite{mehrbakhsh2023adversarial} in the context of image recognition. As another example, \cite{srivastava2022beyond, mizrahi2023state} observed a class of LLMs being given different relative rankings based on a choice between semantically equivalent prompting templates. For a fluent English speaker, the prompt templates appear to be roughly equally straightforward, providing no reason to believe that performance ranks should differ based on template choice.  In short, unknown characteristics of LLMs or benchmark datasets can affect their relative performance in unpredictable ways.

In our example, the scope of the ``format-following'' task is not well-defined. Even if it were, the unexpected sensitivity of rankings to certain characteristics of the benchmark dataset means that rank reversals are likely when these characteristics are not accounted for. Ideally, we would have a clear scope for the task and account for uncertainties associated with the selection of evaluation items, variability in evaluator scores, variability in model responses, and so forth.

\begin{table}[]
\centering
\begin{tabular}{@{}lcc@{}}
\toprule
                                                      & \multicolumn{2}{c}{\textbf{Success Rate}} \\ \cmidrule(l){2-3} 
                                                      & Easy questions     & Hard questions     \\ \midrule
\multicolumn{1}{r}{\textbf{Llama 2 7b}}               & \textbf{88\%}      & 36\%                 \\
\multicolumn{1}{r}{\textbf{Mistral 7b instruct v0.1}} & 80\%               & \textbf{45\%}        \\ \bottomrule
\end{tabular}
\caption{Example of a rank reversal between two models on the Format-Following benchmark dataset. We estimated question difficulty based on the average performance of a class of LLMs, leading to the following equally sized categories of questions: easy, hard, and expert. We ignore the expert category as the performance of open-source models is very low for these questions.}
\label{tab:llm-rankings}
\end{table}

\section{Our Proposal: Better-Defined Targets of Estimation Through the Estimands Framework}\label{sec:estimand-framework}

We now describe the estimands framework adapted from \cite{ICH2019} to improve the quality of inference and reporting in ML evaluations. The framework provides a structure for ML evaluations that emphasizes and clearly defines the target of estimation, i.e., the estimand. As we will see, precisely defining an estimand is a multi-step process that is more subtle than might appear at first glance. For instance, the counterfactual or potential outcome frameworks in causal inference are centered around defining the causal estimand \citep{hofler2005causal}. \cite{ICH2019} discusses many considerations that affect the treatment effect estimand in clinical trials, from the definition of the target population, to the handling of intercurrent events\footnote{Intercurrent events are ``events during the study that may complicate the definition and estimation of the treatment effect estimand, such as premature discontinuation of randomized treatment, taking rescue medication, or death'' \citep{darken2020attributable}}. We discuss analogous issues in ML evaluation.

\subsection{The Estimands Framework}\label{sec:estimands-framework-detailed-definition}

We describe our proposed estimands framework as a series of steps to take in evaluations. Example applications are given next in Section \ref{sec:application_estimands_framework}.

\subsubsection{Define the Evaluation Objective and Subject}

The first step is to identify an evaluation objective and the subject of evaluation. The evaluation objective is the purpose of the evaluation, and the subject of evaluation can be one or more training algorithms, trained models, or machine learning systems.

\subsubsection{Define the Estimand}

The second step is to translate the evaluation objective into a precisely defined estimand. Defining an estimand requires a description of the following characteristics (see Figure \ref{fig:estimand-framework}):
\begin{enumerate}[(A)]
    \item The \textbf{scope or population} of interest that the evaluation aims to cover, i.e. the context of application of an ML model. Clearly defining a scope or population of interest can be challenging. We discuss this further in Section \ref{sec:llm-example-review} in application to LLM evaluation.
    \item A \textbf{data acquisition/handling} strategy, describing how evaluation data is obtained and how data issues are handled. For example, if data annotators are employed, how are inconsistencies in labeling or missing labels handled? How reliable are the labels and are labeling errors addressed? How are dependencies between data points (e.g., temporal dependencies) accounted for? How representative is the evaluation data for the give population? Choices in the acquisition or handling of evaluation data impact what is being estimated and, consequently, how evaluation results should be interpreted and applied for decision-making. We discuss an example of the impact of these considerations in Section \ref{sec:clustering-answer} in application to clustering evaluation.
    \item A choice of \textbf{metric} (or metrics) to measure elements of the scope/population, such as squared error for regression problems.
    \item  A choice of \textbf{aggregation} method to summarize the behavior of the metric (A) over its scope (B), such as an arithmetic or geometric mean. When considering multiple metrics or measurements, the choice of aggregation procedure also concerns how they should be combined into an overall summary.
\end{enumerate}

\subsubsection{Choose Estimation Methodology}

The third step is to choose an estimation methodology (in short, a main estimator) for the task at hand. As we show next in Section \ref{sec:application_estimands_framework}, the appropriateness of a given estimation methodology depends on characteristics of the estimand and of evaluation data. Given evaluation data, the main estimator produces a \textbf{main estimate}, e.g., a numerical value that summarizes a model's performance.

\subsubsection{Choose Uncertainty Estimation Methodology}

The fourth step is to choose a method to estimate uncertainty (in short, an uncertainty estimator to provide a confidence interval) regarding the estimand. The goal is to account for uncertainty in the numerical value of the estimate that can be due to small sample sizes, variability in the evaluation data, labeling uncertainty, or uncertainty regarding assumptions that underlie the choice of evaluation methodology. In some cases, the main estimate and its uncertainty can be jointly estimated. In other cases, uncertainty may need to be characterized based on sensitivity analyses or other techniques. For example, one can account for sensitivity to estimation methods by reporting the range of estimates from different methodologies.

\subsubsection{Reporting Standard}

Given an evaluation that follows these steps, an external observer should be able to investigate whether or not the choice of estimand aligns with the evaluation objective, and whether the estimation methodology is adequate. They should also understand the scope of applicability of the evaluation's results and the extent to which results can transfer to other applications. For this, the components of the estimands framework should be put forward rather than relegated to footnotes and caveats.

\subsection{Application of the Estimands Framework to Rank Reversal Examples} \label{sec:application_estimands_framework}

In this section, we walk through each of our rank reversal examples and apply our estimands framework from Section \ref{sec:estimand-framework} to describe the target of estimation. Then, we show how the estimands framework helps explain the cause of the observed rank reversals and suggest better or alternative evaluation approaches. We discuss incorporating data acquisition and handling issues in the definition of an estimand in Section \ref{sec:data_acquisition_handling_example}. In Section \ref{sec:MCDM}, we discuss the use of multi-criteria decision-making methods (MCDM) to aggregate multiple scores into one and to help report on nuanced evaluation results.

Importantly, note that the contributions of this section are not the solutions and best practices that we mention. {Rather, our goal is to show how the estimands framework can be used to systematically guide the practice and review of ML evaluations.}

\subsubsection{California Housing Dataset}\label{sec:cross-validation-answer}

Here are the components of the estimands framework given in the California Housing Dataset example of Section \ref{sec:california_example}.

The evaluation objective is to estimate the expected generalization error of two models, with the ultimate goal of selecting the better model. We consider as subjects of evaluation a linear model and a decision tree model trained on a given dataset $\mathcal{D}_n$, $n=2,000$.

The estimand is defined as follows. Our target population is all California Census block groups. The evaluation data is the training dataset, and other data acquisition/handling issues are ignored. The metric is the squared error, and we aggregate over the population using the mean. 

As an estimator, we consider a leave-one-out cross-validation with no uncertainty quantification.

\paragraph{Cause of the Rank Reversals}

We can identify the cause for the high probability of rank reversal by walking through the characteristics of the estimand and the models.

\begin{figure}[!ht]
    \centering
    \includegraphics{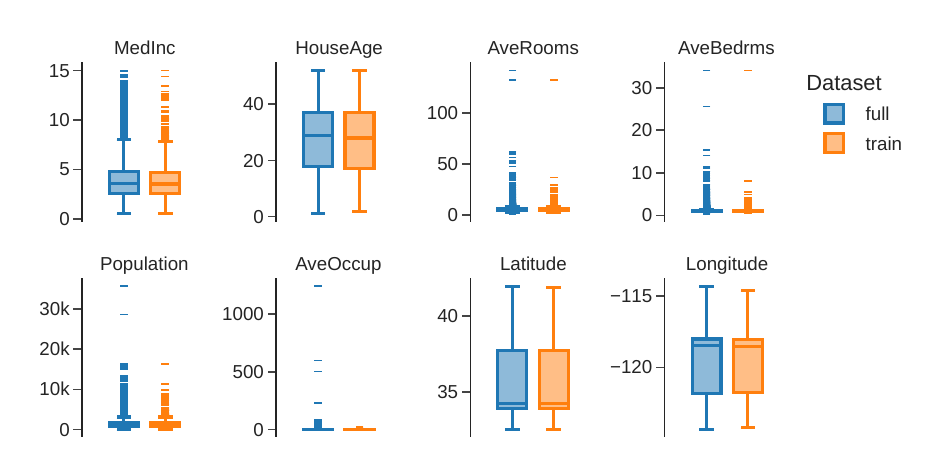}
    \caption{Comparing the distribution of features between the full California Housing Dataset and a training dataset of $2,000$ random examples. Notice the four heavy-tailed features: the average number of rooms, the average number of bedrooms, the block group population, and the average occupation. Census block groups with outlying average occupation numbers are not represented in the training dataset.}
    \label{fig:california-features}
\end{figure}

Regarding the target population, we have access to features for all California Census block groups. This can be used to verify that the evaluation data is representative of all block groups, or assess problems that can arise in predictions related to unrepresented subgroups. Figure \ref{fig:california-features} compares the distribution of features in an evaluation dataset to the full California dataset. We can see that many features have heavy-tailed distributions, and that block groups with outlying average occupation are not represented in the evaluation data. This is a problem given that the squared error metric and mean aggregation are both sensitive to extreme values.

Thinking about the models under consideration, we cannot expect a linear model to generalize well to outlying out-of-sample block groups, unless the linearity assumption was thoroughly justified and checked. In fact, in our example, we have median house price predictions in the \textit{negative} millions for some block groups that are not in the training data. These extremely inaccurate predictions are not reflected in cross-validation estimates of the linear model, leading to a high probability of rank reversal.

In short, we see that the cross-validation estimator is not suitable for the given estimand, given unrepresented outlying districts from the target population, the squared error metric, and the mean aggregation.



\paragraph{Potential Solutions to the Rank Reversals}

There are multiple ways to address the rank reversal problem that we can identify by walking through the steps of the estimands framework and making appropriate changes.

First, we can question whether the choice of estimand is appropriate for the evaluation objective. This is highly application-specific but, given the heavy-tailed distribution of some block group features, we may prefer an estimand that provides more detailed insight into the behavior of models. For instance, we could separate city block groups from suburban and rural block groups, and aim to separately estimate generalization performance for each of these subgroups. If appropriate for a given application, alternative metrics and aggregations can be selected.

Second, assuming we keep our original estimand, we can question whether the training dataset was suitably selected to represent our target population. Given the existence of block groups with outlying features, a weighted sampling or stratified sampling scheme may be preferable to random sampling.

Third, we can rethink our estimation methodology. Given our model selection goal, we should complement any performance estimator with additional considerations that help inform an appropriate model selection decision. We can question the validity of assumptions that underlie our models, such as the linearity assumption. We can also check the validity of predictions on outlying block groups that are not in the training data. Specifically, we know that median house prices are positive and we should be able to provide a plausible upper-bound based on subject-matter expertise. This can be used to check whether a model is generalizing properly to the full population of Census block groups. Alternatives to the cross-validation estimator may also be considered, such as confidence-based performance estimation \citep{bialek2024we}, direct loss estimation \citep{NannyML2024}, or performance estimation through assessor models \citep{hernandez2022training}. These methods rely on features of the entire population, not just on features of the training dataset.

Finally, we may want to quantify uncertainty regarding cross-validation estimates in order to better understand the informativeness of these estimates for model selection. Unfortunately, methods that only rely on a training dataset (such as the cross-validation confidence interval method of \cite{bates2023cross}) are not suitable here. Information about block groups with outlying features needs to be accounted for in order to detect the extrapolation problem associated with the linear regression model. Sensitivity analyses and simulation studies would be useful to characterize the accuracy of the cross-validation estimators in this example.

\subsubsection{Clustering Evaluation} \label{sec:clustering-answer}

Here are the estimands framework components for the clustering evaluation example of  Section \ref{sec:clustering-example}.

The evaluation objective is to determine which of two models performs better to cluster a set of 400 face images that belong to 40 individuals. The metric is a binary indicator of whether a pair of two images represent the same individual or not, i.e., whether they match or not. That is, for two images $i$ and $j$, we write $c_{i,j}=1$ when $i$ and $j$ are a match and $c_{i,j} = 0$ otherwise, and we write $\hat c_{i,j}$ for the model prediction. This is then aggregated into an F-score, the harmonic mean of precision and recall. The scope or population of interest is the set of all ${400 \choose 2}$ image pairs $(i,j)$. Formally, the F-score that we want to estimate can be written as
\begin{equation}
    F = \frac{\sum_{i,j} c_{i,j} \hat c_{i,j}}{\left(\sum_{i,j} c_{i,j} + \sum_{i,j} \hat c_{i,j}\right)/2}.
\end{equation}
The data acquisition strategy is to sample 10 clusters at random and to obtain the matching status of all pairs among them. We discuss practical data acquisition issues in more detail in a subsection below.

\paragraph{Cause for Rank Reversals}

Going through the components of the estimands framework, the cause for rank reversals quickly becomes apparent.

Considering the target population (the full dataset), when sampling 10 clusters at random, the ratio of matching to non-matching pairs in the evaluation data is not representative of the full dataset. Indeed, in the full dataset of 400 images, there are ${10 \choose 2} = 45$ matching pairs for each of the 40 individuals, and $78,000$ non-matching pairs in total, resulting in a match to non-match ratio of $1,800/78,000 \approx 2.3\%$. In a sample of 10 random clusters, there are $450$ matching pairs, and $4500$ non-matching pairs, resulting in a matching ratio of $10\%$.

The difference in the distribution of classes creates a bias in precision estimates, but not in recall estimates. As such, rank reversals in for the aggregate F-score can occur when comparing models that do not have the same recall. The problem is studied extensively in \cite{foxcroft2024class}.

\paragraph{Solution to the Rank Reversals}

A solution from \cite{binette2023estimating, binette2024evaluate} is to change how the estimand is described in order to facilitate its estimation from a cluster sample. To do so, the F-score is expressed as a function of cluster metrics rather than pairwise errors, and this representation is then used to derive an estimator that is unbiased for random cluster sampling. That is, we consider the population of clusters, rather than the population of pairs of images, and we change the metric and aggregation to still obtain the same F-score as a result. A random sample of clusters is then representative of the population and accurate estimates can be obtained from the estimators described in \cite{binette2024evaluate}.


\paragraph{Practical Data Acquisition and Handling Issues}\label{sec:data_acquisition_handling_example}

In practice, we often need data annotators to label data, here to identify ``ground truth'' clusters. This is a difficult task when true cluster membership cannot be verified exactly. As such, biases, inaccuracies, and inconsistencies in the data labeling, and how they are handled, affect what is being estimated.

For example, in \cite{binette2023estimating}, we evaluated a large-scale identity clustering system for patent inventors based on a probabilistic random sample of 400 ground truth clusters identified by a team of data annotators. In order to facilitate the process, data annotators could access the system's current clustering predictions and use them as a starting point of the labeling, obtaining true clusters by cleaning and merging predicted clusters as needed. This approach created a bias towards current predictions, since they were used as a default. 

In other words, we were not estimating the ``true'' F-score of the system. We were estimating the difference (summarized by a F-score) between current predictions and a manually corrected version of these predictions, where only sufficiently obvious errors were accounted for. This estimand that should have been directly stated, instead of mentioning biases as a potential caveat of the estimation methodology. This estimand that accounts for data acquisition issues is clear, unambiguous, and still perfectly relevant to the objective of improving predictions.

In short, we believe that an estimand needs to account for data acquisition and handling issues in order to be well-defined. Even though we would ideally want to estimate $\varphi$, if data acquisition and handling issues cause us to estimate $\varphi'$, then the latter needs to be stated as the estimand. This way, the results of an evaluation can be properly interpreted without having to analyze in detail the estimation methodology, its assumptions, and its caveats. 

\subsubsection{LLM Evaluation}\label{sec:llm-example-review}

Here are the components of the estimands framework given in the example of Section \ref{sec:LLM-rank-reversal}. Our goal is to determine which of two LLMs is better at following a wide range of formatting instructions. Our metric is a binary indicator of success on evaluation items, and this is aggregated into an unweighted average success rate. The scope of formatting instructions is implicitly defined through the Format-Following benchmark dataset. Data acquisition and handling issues are ignored.

\paragraph{Cause of Rank Reversals}

The estimand is too vaguely defined for the evaluation objective, since it depends on important but unmeasured characteristics of the benchmark dataset. Relatively small variations in the composition of the benchmark dataset can lead to unexpected rank reversals, due to the fact that the mean aggregation does not account for characteristics of the data. Without doing an in-depth analysis of the benchmark dataset, we don't know what the rankings represent, and we don't know to what kind of format-following tasks they would generalize.

\paragraph{What Is a Solution?}

A solution is to better define the estimand through the specification of a clear scope for the format-following task. For instance, a probability distribution of format-following prompts can be specified through prompt templates or a generative prompt model. This can clarify the composition of the benchmark dataset at a high level and help understand the applicability of evaluation results. This aligns with the following recommendation from \cite{davis2023benchmarks}:

\begin{quote}``Benchmark sets should be constructed using a well-defined and replicable methodology. If one considers the process of running some AI system $S$ on benchmark $B$ to be an experiment testing a hypothesis, then the hypothesis `System $S$ achieves performance $P$ over problems with characteristics $X$, $Y$, $Z$,' is a considerably more meaningful and cogent statement than `System $S$ achieves performance $P$ on the specific benchmark $B$.' If $B$ can be claimed to be a representative or a random selection of problems with characteristics $S$, $Y$, $Z$ then there is some support for the stronger statement. If there is only the benchmark set $B$ constructed catch-as-catch-can, then it is hard to know how these results will generalize. In practice, this is rare, except for benchmarks created using automatic synthesis."
\end{quote}

Now, it could be difficult to define a single scope that is representative a broad range of applications. Instead, we can evaluate on multiple narrow scopes, and then aggregate the results through multi-criteria decision-making (MCDM) methods (see next section). For example, we can separately evaluate performance for different format types (e.g. JSON, markdown), for different prompting templates, at different difficulty levels, and so forth. If a single model performs best across all tasks, it can be declared best. Otherwise, we can use MCDM methods to help provide a single score aggregate and detailed evaluation results that are useful for a variety of applications.

More refined approaches to the definition of an estimand might use pyschometric methods. For instance, \cite{mlayoutstutorial2024} introduces ``measurement layouts,'' a framework to characterize system performance through the relationship between system capabilities and the characteristics of evaluation items \citep{mlayoutstutorial2024}. Key in this framework is the estimation of clearly scoped capabilities through the use of evaluation item characteristics. The estimated capability characteristics of a system can then be used to predict performance on new items.

\paragraph{Multi-Criteria Decision-Making and the Pareto Frontier}\label{sec:MCDM}

Multi-criteria decision-making (MCDM) \citep{triantaphyllou1998multi, greco2016multiple} is the task of identifying an optimal solution to a problem given multiple criteria that can conflict with one another. For example, we may want to choose an LLM that balances performance across many different tasks. Often, no single solution is better than alternatives for all criteria, resulting in a set of non-dominated solutions called the Pareto frontier (see \cite{martinez2018between, RajiBuolamwini2024} for the use the Pareto Frontier in AI evaluation).

The Pareto frontier can be easily visualized when two or three criteria are considered. When more criteria are considered, dimensionality reduction and data visualization techniques can be used to summarize the behavior of alternatives across criteria \citep{ibrahim20163d}.

Another focus of MCDM is the translation of subjective preferences into an acceptable decision. For instance, the Analytic Hierarchy Process (AHP) \citep{saaty2003decision} translates pairwise relative preferences between criteria into a weighting scheme for computing a weighted average. There is never a single ``true'' or ``best'' solution to a choice of score aggregation. This is always a subjective process. The key is for the process to be transparent and for useful information to be provided.

\section{Discussion}\label{sec:discussion_estimands_framework}

We highlighted current issues in the evaluation of ML models: (1) validity issues due to a lack of consideration of inference and reporting as key components of evaluation, beyond merely computing metrics, and (2) a high probability of rank reversals in specific cross-validation, clustering evaluation, and LLM evaluation applications. To address these problems, we proposed emphasizing the role of estimands (the targets of estimation) in ML evaluation, applying the estimands framework from clinical trials biostatistics that formalizes essential components of experimental design. We showcased how defining estimands through four key components helps identify methodological problems, highlight appropriate estimation methodologies, and properly interpret evaluation results.

Using an estimands framework to clearly define targets of estimation and to structure evaluations unlocks a number of key benefits. First, estimands enable the use of more sophisticated statistical methodologies to reduce the cost of evaluations. For instance, scoring LLMs on a large set of questions can be expensive. Through an estimand that specifies a clear population of questions, we can efficiently sample a subset that provides a sufficiently accurate performance estimate at a lower cost. Second, well-defined estimands enable uncertainty quantification. Without an estimand, we are simply computing metrics for which there is no context for uncertainty. An estimand provides a meaningful target of inference regarding which probability statements can be made. Finally, requirements for the definition of an estimand and for the structure of evaluations can be used in AI governance and AI auditing to help verify the quality and relevance of evaluations.


\section*{Acknowledgements}

The authors would like to thank David Banks, Giri Gopalan, Santiago Viquez, and José Hernández-Orallo for sharing useful comments and perspectives. The authors would like to acknowledge the US Department of Energy, National Nuclear Security Administration’s Office of Defense Nuclear Nonproliferation Research and Development (NA-22) for supporting this work. Paper approved for public release. LA-UR-24-25563.

\bibliographystyle{chicago}
\bibliography{biblio}

\end{document}